\documentclass[conference]{IEEEtran}
\IEEEoverridecommandlockouts
\usepackage{cite}
\usepackage{amsmath,amssymb,amsfonts}
\usepackage{algorithmic}
\usepackage{graphicx}
\usepackage{textcomp}
\usepackage{xcolor}
\usepackage{subcaption}
\def\BibTeX{{\rm B\kern-.05em{\sc i\kern-.025em b}\kern-.08em
    T\kern-.1667em\lower.7ex\hbox{E}\kern-.125emX}}
\begin{document}

\title{A simple vision-based navigation and control strategy for autonomous drone racing \\
\thanks{The work presented in this paper was supported by the AGH University of Science and Technology project no. 16.16.120.773}
}

\author{\IEEEauthorblockN{Artur Cyba}
		\IEEEauthorblockA{
			AGH University of Science \\
			and Technology, Krakow, Poland \\
			E-mail: arturcyba@student.agh.edu.pl}
		\and
		\IEEEauthorblockN{Hubert Szolc}
		\IEEEauthorblockA{
			AGH University of Science \\
			and Technology, Krakow, Poland \\
			E-mail: szolc@agh.edu.pl}
		\and
		\IEEEauthorblockN{Tomasz Kryjak, \textit{Senior Member IEEE}}
		\IEEEauthorblockA{ AGH University of Science \\
			and Technology, Krakow, Poland \\
			E-mail: tomasz.kryjak@agh.edu.pl}
}

\maketitle

\begin{abstract}

In this paper, we present a control system that allows a drone to fly autonomously through a series of gates marked with ArUco tags.
A simple and low-cost DJI Tello EDU quad-rotor platform was used.
Based on the API provided by the manufacturer, we have created a Python application that enables the communication with the drone over WiFi, realises drone positioning based on visual feedback, and generates control.
Two control strategies were proposed, compared, and critically analysed.
In addition, the accuracy of the positioning method used was measured.
The application was evaluated on a laptop computer (about 40 fps) and a Nvidia Jetson TX2 embedded GPU platform (about 25 fps).
We provide the developed code on GitHub. 


\end{abstract}

\begin{IEEEkeywords}
drone racing, AruCo markers, Tello, OpenCV, autonomous drones
\end{IEEEkeywords}

\section{Introduction}

During the last few years, we have seen a significant increase in interest in unmanned aerial vehicles (UAV).
This is especially true for various autonomous operations --  e.g., delivery of various types of goods, surveillance, inspection of technological facilities (power lines), intelligent agriculture, or search for people, as well as military applications.
One interesting application is also autonomous drone\footnote{In this paper we deliberately narrow the term ``drone'' to a~multi-rotor (like quad-copter) UAV}  racing.
In~these, the vehicle must fly along a track denoted by a set of~gates of~specified shape and~colour in~the shortest possible time.
The~popularity of this type of competition is evidenced by the many challenges held in~last years -- e.g., AlphaPilot -- Lockheed Martin AI Drone Racing (2019), IROS Autonomous Drone Racing (2016--2019), and Game of Drones: A NeurIPS Competition (2019).
In~the first two cases, the competition took place in~the real world, while in~the last one -- virtually, with~the use of a~simulator.
Most of the teams participating in the aforementioned competitions base their solutions on complex control algorithms.
They use not only vision information, but also fuse data from other sensors, such as the IMU (Inertial Measurement Unit).
It should be noted that although the races themselves have no practical commercial application and are rather a~scientific and engineering challenge, as well as simply good fun, the developed vision and trajectory generation and realisation methods can be applied e.g. in the fast search of closed spaces.


In this project, we  propose a~simple but still effective algorithm that performs a~flight through a track denoted by a~series of gates. 
Our intention was to prepare a~basis from which one can easily start research in the~area of autonomous drone racing.
For this reason, we decided to use the DJI Tello EDU drone because it is relatively cheap, affordable, and safe.
It allows both manual and~autonomous control from~the~level of software running on a~ground~station, such as a~laptop or an Nvidia Jetson development board.
Unfortunately, one of its major drawbacks is the control of only $(v_x,v_y,v_z,\omega_z)$ velocity vector, while custom drones allow to directly control of $(roll, pitch, yaw)$ angles.
The operation and settings of the internal controller, which enables the drone to achieve the set velocity vector, are implemented by the manufacturer and~not directly available to the user.


The main contributions of this work are two original drone control strategies for autonomous drone racing, with particular emphasis on the AlphaPilot competition.
They only use visual feedback to generate the flight trajectory.
The aforementioned algorithms have been implemented on a~general purpose computer and an embedded GPU platform.
The results of the second experiment showed that the calculations could be carried out online on the drone, without the need to communicate with the ground station.
All codes developed by us are available in our GitHub repository\footnote{https://github.com/vision-agh/drone\_racing\_tello/} for use in other projects, including classes with students.
We also provide an instruction on how to start the system on a~PC and Jetson TX2 platform, as well as a~video abstract in which we present an example of a~flight through a~series of gates. 


The reminder of this paper is organised as follows.
Section \ref{sec:related work} provides an overview of previously published work related to~autonomous drone racing.
In~Section \ref{sec:the proposed system} we describe our solution, including the two proposed control strategies.
We present the evaluation of the system in Section \ref{sec:exp results}.
Section \ref{sec:conclusions} summarises our work and~describes plans for future project development.

\section{Related work}
\label{sec:related work}

Most of the algorithms proposed in~scientific literature to realise autonomous drone flight through gates can be divided into two parts.
The first one is responsible for gate detection based on visual information.
The purpose of the second is to properly control the drone. 
The fusion of data from the camera and other sensors, such as IMU, is often used for this purpose.


\subsection{Gate detection}

\begin{figure}
    \begin{minipage}{0.50\textwidth}
     \begin{subfigure}[b]{0.31\textwidth}
         \centering
         \includegraphics[width=\textwidth]{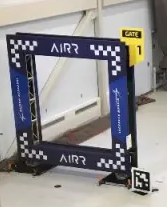}
         \caption{}
         \label{fig:gate alpha pilot}
     \end{subfigure}
     \hfill
     \begin{subfigure}[b]{0.31\textwidth}
         \centering
         \includegraphics[width=\textwidth]{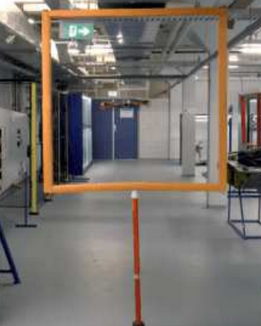}
         \caption{}
         \label{fig:gate iros}
     \end{subfigure}
     \hfill
     \begin{subfigure}[b]{0.31\textwidth}
         \centering
         \includegraphics[width=\textwidth]{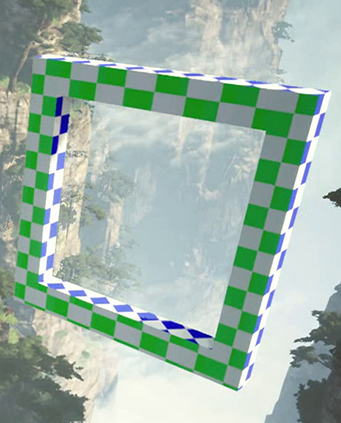}
         \caption{}
         \label{fig:gate game of drones}
     \end{subfigure}
        
        \caption{Gates used in different autonomous drone racing contests: (a) Alpha Pilot\textsuperscript{a}, (b) IROS \cite{Li2018} and (c) Game of Drones\textsuperscript{b}}
        
        \textsuperscript{a}\tiny{https://www.techbriefs.com/component/content/article/tb/webcasts/podcasts/35575} \\ 
    	\textsuperscript{b}\tiny{https://www.microsoft.com/en-us/research/blog/game-of-drones-at-neurips-2019-simulation-based-drone-racing-competition-built-on-airsim/?OCID=msr\_blog\_gameofdrones\_neurips\_fb}
        \label{fig:gates}
    
    \end{minipage}
\end{figure}

As we have pointed out in the introduction, many autonomous drone races have been organised in the last years.
During each of them, different designs of the gates were used.
We present a~selection of them in Figure \ref{fig:gates}.
Currently, deep convolutional neural networks (DCNN) are most commonly used for their detection.
They are often specifically designed for this particular task.


In paper \cite{Cocoma-ortega2019} a~modification of the PosNet network that allows the position $(x, y, z)$ of a drone to be determined relative to the centre of a~detected gate was proposed.
The authors tested the solution in simulation and in a real environment, obtaining a~maximum processing rate of 100 FPS on a GPU.
Gates from IROS Autonomous Drone Race were used in this project.


A~similar approach was presented in the paper \cite{Jung2018}.
The authors proposed a~new network for gate detection, called ADRNet.
It was created by AlexNet modifications, which included removing some layers that were irrelevant according to the authors.
This allowed them to achieve processing at 30 FPS on a GPU.
The authors also used gates from the IROS Autonomous Drone Race and conducted successful tests in a~real environment.


A~neural network for gate detection was also used in the work \cite{Kim2020}.
The authors trained MobileNetSSD for this purpose using augmented data collected in a~simulator.
Bounding boxes were then determined for all gates detected with probability above 90\%.
From these, one (with the largest area) was selected as the location of the nearest gate.
The algorithm was successfully tested in the Game of Drones competition at NeurIPS 2019.


Another example of using an off-the-shelf neural network architecture for gate detection is the work \cite{Foehn2020}.
In this case, the authors used a~5-level U-Net that detected the four vertices of all square frames in the field of view of the camera.
To train the network, 28000 images collected in 5~different environments were used.
Inference took place on an Nvidia Jetson Xavier using half precision (FP16).
The algorithm was successfully tested in a~real environment during the AlphaPilot 2019 competition.


A~different approach was presented in the paper \cite{Li2020a}.
The authors developed a~novel Snake Gate Detection algorithm.
It is based on the colour difference between pairs of compared points.
As a~result, its effectiveness depends primarily on the chosen difference threshold and the gate design.
In this case, the authors used homogeneous orange gates from the IROS Autonomous Drone Racing.
This allowed the algorithm to make correct detections, which was verified in a~real environment.


\subsection{Vision--based drone control}

Many algorithms with different degrees of complexity have been proposed for controlling a~drone while flying through a series of gates.
One of the simpler methods is presented in the paper \cite{Cocoma-ortega2019}.
It used as the input the  $(x, y, z)$ position relative to the gate obtained from the vision algorithm and the spatial orientation obtained from the IMU.
Based on this, the drone positioned itself centrally relative to the gate and then decreased the distance.
In this way, it reached a~point from which the gate did not fit within the field of view of the camera.
The drone then flew in a~straight line for a~set distance, determined by the last position measurement.
After this, it searched for the next gate and repeated the same steps.
The authors conducted both simulation and real-world tests, which confirmed the effectiveness of the proposed method, with an average positioning error of 20 cm.


A~different approach was proposed in the work \cite{Kim2020}.
In it, three perpendicular linear speeds of the drone and a~yaw axis rotational speed were controlled.
Each of them was treated separately, using a classical PD controller for the three mentioned quantities.
The exception was the velocity perpendicular to the gate plane.
In this case, a P~regulator was used.
It minimised the estimated distance to the gate, which was determined as a linear combination of the current position of the drone and the location of the gate obtained from the vision algorithm.
The proposed solution was tested in a~simulator in the Game of Drones competition at NeurIPS 2019, where it obtained a~second place in two competitions.


Classical proportional controllers were also used in the work \cite{Li2020}.
In this case, they were arranged in a~cascade scheme.
The outer controller was responsible for minimising the position error, while the inner controller was responsible for minimising the velocity error.
The obtained control was fed to the Paparazzi autopilot.
This off-the-shelf module was responsible for controlling the speed of the drone's rotors, again through the built-in cascade of P~regulators.
The authors also proposed a~novel VML (Visual Model-predictive Localisation) algorithm.
It enables a~more accurate (than the classical Kalman filter) estimation of the drone's state by fusing video information with AHRS (Attitude and Heading Reference System) data.
The entire system has been tested in a~real-world environment using a~light-weight Trashcan racing drone, successfully flying through the gate track at an average speed of 2 m/s (maximum 2.6 m/s).


A~different control algorithm was proposed in the work of \cite{Foehn2020}.
It used the Pontryagin's maximum principle for sampling-based receding horizon path planning.
On this basis, a~cascade of controllers was defined: outer for position control and inner for attitude control.
Their operation was based on VIO (Visual-Inertial Odometry).
The authors also applied an EKF (Extended Kalman Filter), which corrected the IMU measurements using the position of all gates detected in the image.
The proposed algorithm was used in the AlphaPilot 2019 competition, where it obtained the second place.
In doing so, it provided a~flight efficiency of 100\% at 5 m/s and 60\% at 8 m/s.


A~slightly different approach was proposed in the work \cite{Jung2018}.
It used the LOS (Line-Of-Sight) guidance algorithm, originally developed for fixed-wing UAVs, to control the quadrotor. 
It was modified by decoupling the lateral and yaw axes and then verified in a~real environment.
The authors achieved a~successful flight through a~track of 9 randomly placed gates.


A~completely different approach was used in the work \cite{Rojas-Perez2020}.
The authors proposed to combine the task of gate detection and drone control in the form of a~single convolutional DeepPilot network.
It was developed based on PosNet and consists of three parallel branches.
Each branch is responsible for determining specific control commands, respectively: 2D orientation (roll and pitch angles), yaw rotation speed, and vertical linear speed.
Simulation data with reference control commands collected during manual flights were used to train the network.
The proposed algorithm was tested only in a~simulation environment on tracks consisting of a~different number of gates placed in different configurations.
Successful trajectory execution at an average processing frequency of 25 fps was obtained.


It is worth noting that autonomous drone racing is a~relatively new topic present in scientific works.
Indeed, all presented papers were published between 2018--2020.
Most algorithms use neural networks for gate detection (regardless of their design) and data fusion from multiple sensors to control the drone.
They are thus computationally demanding and need a~considerable amount of training data.


\section{The proposed system}
\label{sec:the proposed system}

In this section, the used platform, the vision-based drone position estimation method, and the two proposed control strategies are presented.


\subsection{Drone specification}

In this project, we have used the DJI Tello EDU drone.
It is a light-weight (92g with protective casing) and small (170 $\times$ 170 $\times$ 100 mm)  quadcopter, which is presented in  Figure \ref{fig:tello}.



The drone is equipped with the following sensors: IMU, barometer, downward vision sensor, ToF (Time of Flight) distance sensor and~a 5~Mpix FPV (First Person View) camera.
The communication with the~vehicle is via WiFi and~UDP protocol.
In~this way, by sending special commands (defined by the manufacturer), we can force specific actions and read the data collected by the sensors (including video frames). 
To enable convenient communication with the drone, receiving video data and control, a library in the Python programming language was created, which is based on the requirements and assumptions defined by the manufacturer in the documentation (code available on our GitHub).
The control of the vehicle's movement consists primarily in the presetting of a 3-dimensional vector of linear velocities ($v_x$, $v_y$, $v_z$) and the rotation speed around the yaw axis ($\omega_z$).
In addition to this, it is also possible to execute predefined trajectories, such as flight along an arc or along a straight line for a given distance.


\subsection{Pose estimation}
\label{subsec:pose estimation}


\begin{figure}
    \begin{minipage}{0.45\textwidth}
     \centering
     \begin{subfigure}[b]{0.59\textwidth}
         \centering
         \includegraphics[width=\textwidth]{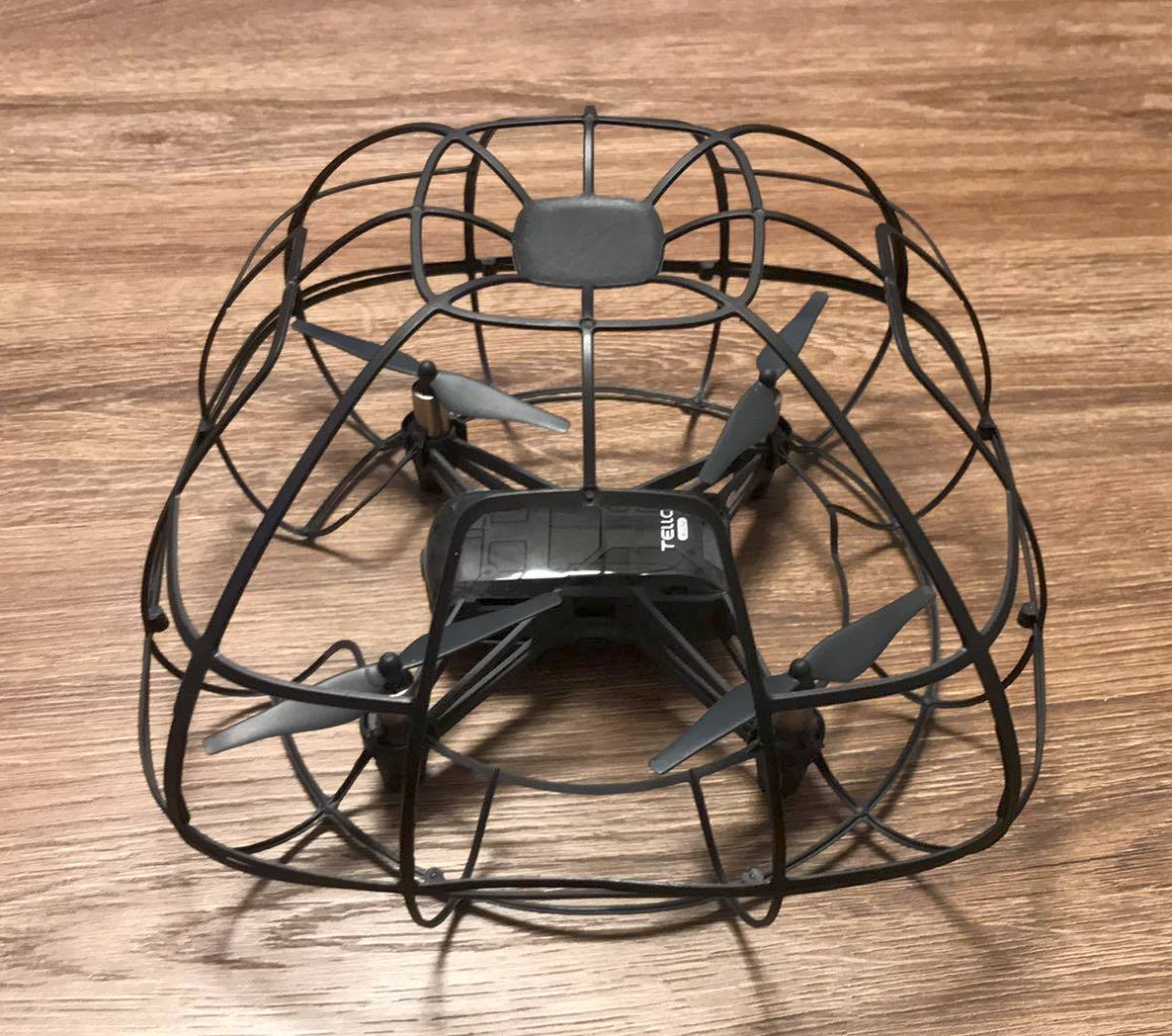}
         \caption{}
         \label{fig:tello}
     \end{subfigure}
     \hfill
     \begin{subfigure}[b]{0.39\textwidth}
         \centering
         \includegraphics[width=\textwidth]{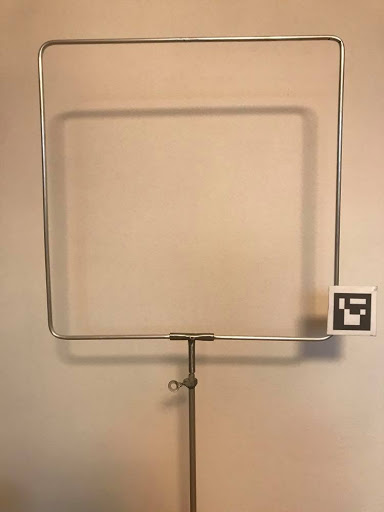}
         \caption{}
         \label{fig:sample gate}
     \end{subfigure}
        \caption{The DJI Tello drone (a) and one of the gates we used in our project (size 50 cm $\times$ 50 cm) (b).}
        \label{fig:tello_gate}
    \end{minipage}
\end{figure}

\begin{figure}[!t]
\centerline{\includegraphics[width=0.4\textwidth]{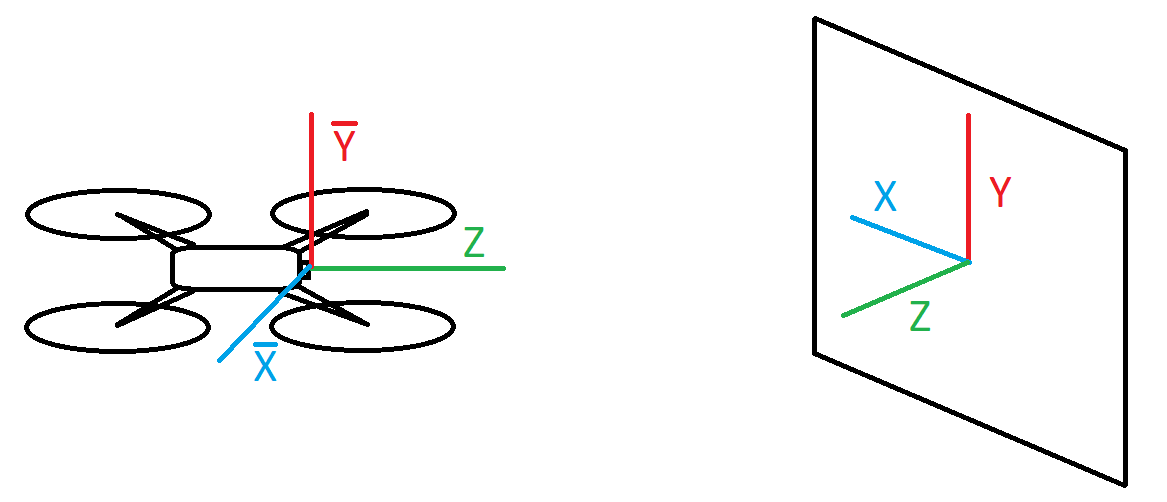}}
\caption{Coordinate system of drone and marker/gate.}
\label{fig:coordinate_system}
\end{figure}

As we mentioned earlier, one of the motivations for our work is the desire to participate in autonomous drone competitions such as AlphaPilot or held in conjunction with the IROS conference.
At this stage, we have decided to use gates with an ArUco  marker \cite{Garrido_Jurado_2014} placed in the bottom right corner, similar to the AlphaPilot competition.
We present one of them in Figure \ref{fig:sample gate}.
These codes allow a relatively accurate estimation of the drone's position relative to the gate.
The system therefore has two coordinate systems -- one associated with the drone and one with the marker/gate.
This is presented in Figure \ref{fig:coordinate_system}.


In the first step, to be able to accurately determine the position of the drone in space using vision data, camera parameters such as focal length and optical centre were determined and then the camera itself calibrated. 
This allowed to remove radial and tangential distortions. 
A calibration pattern in the form of a chessboard and the API available in the Matlab software were used.
The calibration\footnote{Tutorial: https://docs.opencv.org/master/dc/dbb/tutorial\_py\_calibration.html (last access 09.04.2021)} can also be performed using the OpenCV library \cite{opencv_library}. 


In the second step, we used the ArUco marker detection algorithm available in~the OpenCV library. 
It consists of~the following operations: adaptive thresholding, contour extraction (Suzuki \& Be method), determination of vertices (Ramer-Douglas-Peucker algorithm), filtering of quadrangles and~their verification (reading the marker code) -- function $cv2.aruco.detectMarkers()$
In~this way we determine the vertices of all ArUco markers visible in the image.
We then determine the position of the drone relative to them, solving the PnP (Perspective-n-Point) problem for the available 4~points of each marker.
For this, we also use the functionality of the OpenCV library -- function $cv2.aruco.estimatePoseSingleMarkers()$.
Among the obtained translation vectors, we choose the shortest one (w.r.t. Euclidean norm).
It represents the position of the nearest marker/gate through which the drone should fly.
The evaluation, discussed in Section \ref{sec:exp results}, showed that under suitable lighting conditions the positioning is sufficiently accurate.
Additionally, we created a system that handles a situation when the nearest marker is not detected properly for a short period of time. 
In this case, the last known correct position is used as input to the control algorithm.


\subsection{Control algorithm}

For the autonomous flight through the track marked by gates, we have proposed two control strategies.
Both represent simple but at the same time efficient algorithms based only on vision information (without overt fusion with measurements from IMU or other sensors).
The difference between them is the reference system in~which the flight trajectory is generated.
For the first strategy, the drone's native control coordinates are used.
The second strategy, on the other hand, uses the coordinate system associated with the gate, relative to which the drone is positioned according to the~algorithm described in Subsection \ref{subsec:pose estimation}.


\subsubsection{First strategy -- UAV coordinates}

\begin{figure}[!t]
\centerline{\includegraphics[width=0.35\textwidth]{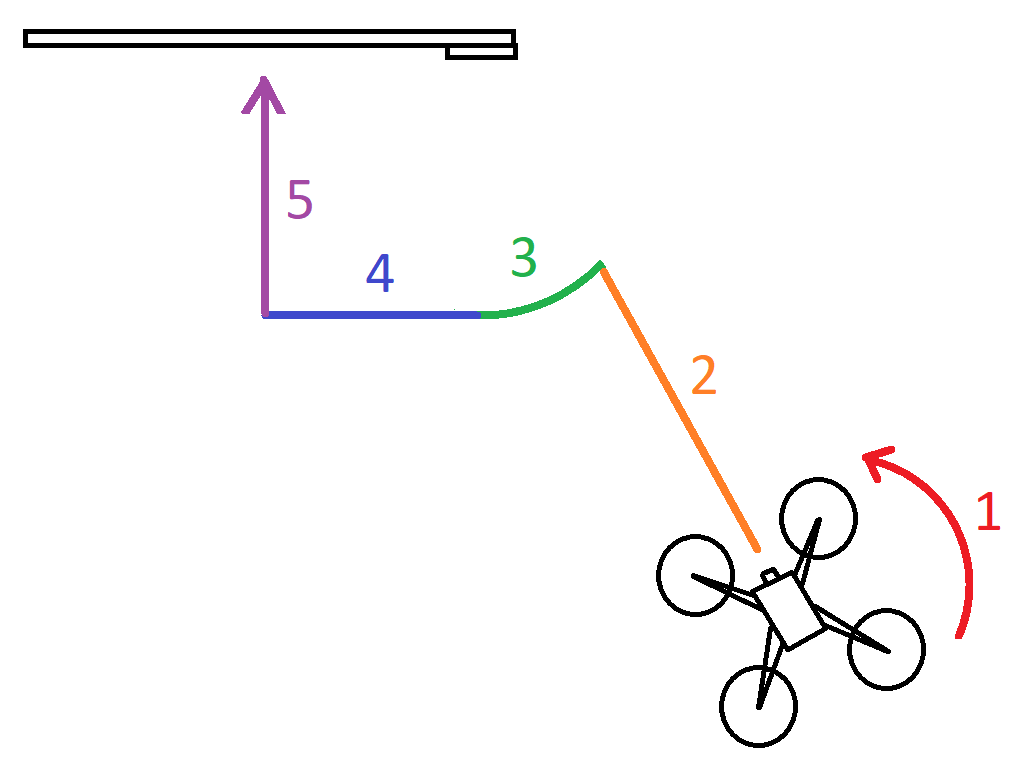}}
\caption{Example trajectory obtained using the first control strategy. It consists of five phases: (1) rotation towards the marker, (2) straight flight towards the marker, (3) facing the plane of the gate, (4) moving towards the centre of the gate, (5) straight flight -- flying through the gate.}
\label{fig:first strategy}
\end{figure}

In the first control strategy, the position and orientation of the vehicle obtained by the vision algorithm are converted to the drone's native control coordinates.
In this way, the vehicle receives information about the location of the gate through which it has to fly.
The generation of the proper trajectory is divided in this case into five phases -- we show a~schematic of an example trajectory in Figure \ref{fig:first strategy}.
At first, the drone positions itself towards the marker (phase 1).
When the angle between the centres of the image and the marker is less than $\alpha_1$, the vehicle starts flying straight ahead (phase 2).
We selected the parameter $\alpha_1 = \tan^{-1}(0.2)$ experimentally as a~compromise between accuracy and smoothness of the motion.
Upon reaching a~distance $d_2$ from the marker, the drone rotates and positions itself facing the gate plane (phase 3).
The parameter $d_2 = 800 [mm]$ is derived from the camera angle and is intended to ensure that the ArUco marker is visible throughout the flight.
Once the drone is facing the gate, it moves parallel to it (phase 4) until it reaches a~point close to the plane perpendicular to the gate and passing through its vertical axis of symmetry.
Then it changes direction again and flies straight ahead (phase 5) until it is on the other side of the gate.
In this case, continuous control of the vehicle's position is not possible because the ArUco marker is no longer within the camera's field of view.
Therefore, this phase lasts for a~predetermined time $t_5 = 5 [s]$.
The value of the parameter $t_5$ is derived from the value $d_2$ and the drone's set speed.
After this time, the drone starts the next iteration of the algorithm, returning to phase 1.

In each phase, a~proportional (P) controller operating with visual feedback ensures that the objectives are met.
We selected the gain values experimentally.
The position of the drone with respect to the vertical axis is also controlled in the same manner.
Thus, during all phases, the vehicle is kept at an altitude that guarantees both the flight over the lower edge of the gate and the visibility of the marker for the visual feedback.


\subsubsection{Second strategy -- ArUco/gate coordinates}

\begin{figure}[!t]
\centerline{\includegraphics[width=0.3\textwidth]{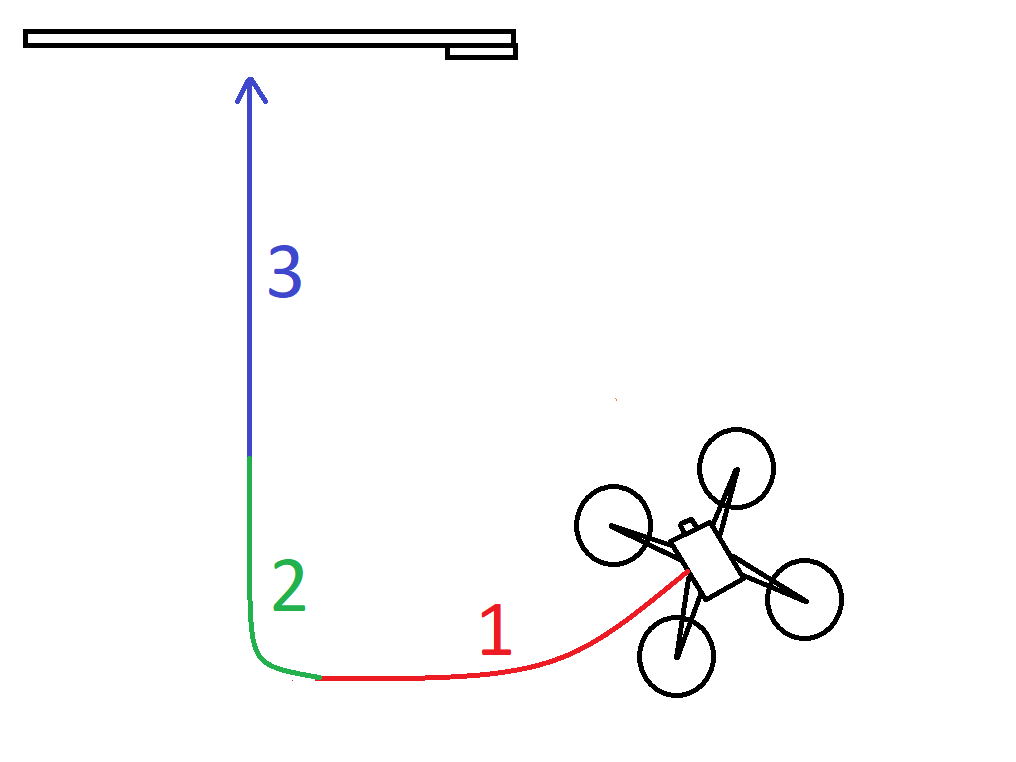}}
\caption{Example trajectory obtained using the second control strategy. It consists of three phases: (1) flight to a point at a certain distance from the gate while keeping the marker in the camera field of view, (2) flight towards the gate, correction of position and orientation relative to the gate if necessary, (3) straight flight -- fly through the gate.}
\label{fig:second strategy}
\end{figure}

We also divided the second strategy into certain phases in which the flight trajectory is generated.
Thanks to the use of ArUco coordinates, in this case their number has been reduced from 5~to 3.
The control is still set in the drone system.
To convert the coordinates, we use simple geometric relationships, unambiguously linking the two systems.
We show a~schematic of an example trajectory possible with this control strategy in Figure \ref{fig:second strategy}.
Initially, the drone flies towards a~point located at a~distance $d_1 = 900 [mm]$ from the centre of the gate at a~suitable altitude relative to the marker. 
It also maintains an orientation pointing the camera towards the centre of the gate.
In this way, phases 1-4 of the first strategy are executed simultaneously.
Obtaining a~distance $d_1$ is necessary to ensure that the ArUco marker remains in the camera's field of view when the drone starts flying directly through the gate.
This provides the vision feedback that is necessary for the correct operation of the P~controller.
When the drone is at a~distance $\delta_2 = 150 [mm]$ from the plane perpendicular to the gate and passing through its vertical axis of symmetry, it enters phase 2 and flies directly towards the gate.
Thanks to the $\delta_2$ parameter, the trajectory is smoothed.
The vehicle does not lose all accumulated kinetic energy, but seamlessly changes the direction of motion.
We selected the value of $\delta_2$ heuristically, based on the parameter $d_1$ and the speed reached by the drone.
Phase 2~is also used to correct the position and orientation of the vehicle before flying straight through the gate.
It occurs when the marker is not visible in the camera image for more than $\Delta t_2 = 0.3 [s]$.
The drone then takes a straight-line trajectory and flies along it for $t_3 = 2 [s]$.
We selected the value of $t_3$ similarly to $t_5$ in the first control strategy.
During this time, the drone flies through the gate and then returns to phase 1~and starts a~new iteration of the algorithm.

As in the first strategy, the realisation of the objectives in each phase is ensured by a~proportional controller operating with visual feedback with gain values selected experimentally.
It is worth noting that phase 1~can theoretically cause the drone to temporarily move away from the gate.
This is necessary due to the vision feedback, which requires the ArUco marker to be in the camera field of view all the time (except for the final fly-through).


\section{Experiments \& results}
\label{sec:exp results}

Our tests focused on three areas.
The first concerned the accuracy of determining the position and orientation of the drone in ArUco coordinates.
This is a~key element due to only using vision-based feedback.
Then, we compared the two proposed control strategies.
We tested their effectiveness when flying through gates in different lighting conditions and measured the time needed to cover the test track.
We conducted these tests using a~computer equipped with an Intel Core i5 processor and an Nvidia GTX 1060 GPU as a~ground station.
Finally, we checked the performance of our system on a~board with an embedded GPU -- Nvidia Jetson TX2.


\subsection{Pose estimation accuracy}

Firstly, we verified the accuracy of the translation vector determination.
We did this by placing the gate with the ArUco marker at random positions relative to the drone.
We collected the reference values manually using a tape measure.
We present the obtained results in Table \ref{tab:translation accuracy}.


\begin{table}[]
\centering
\caption{Selected position accuracy test results.}
\label{tab:translation accuracy}
\begin{tabular}{|c|c|c|c|c|c|c|c|c|}
\hline
\multicolumn{3}{|c|}{Camera pos. [mm]} & \multicolumn{3}{c|}{Reference pos. [mm]} & \multicolumn{3}{c|}{Absolute error [mm]} \\
\hline
x & y & z & x & y & z & x & y & z \\
\hline
145 & -130 & 1106 & 150 & -130 & 1110 & 5 & 0 & 4 \\
-377 & -126 & 800 & -360 & -120 & 820 & 17 & 6 & 20 \\
-119 & -85 & 1564 & -80 & -90 & 1530 & 39 & 5 & 34 \\
-155 & -64 & 825 & -160 & -60 & 830 & 5 & 4 & 5 \\
1 & -8 & 800 & 0 & 0 & 800 & 1 & 8 & 0 \\
10 & -1 & 766 & 0 & 0 & 780 & 10 & 1 & 14 \\
30 & -14 & 851 & 0 & 0 & 840 & 30 & 14 & 11 \\
-7 & -1 & 828 & 0 & 0 & 820 & 7 & 1 & 8 \\
\hline
\multicolumn{6}{|l|}{Mean absolute error [mm] (for all experiments)} & 14.91 & 6.27 & 10.18\\
\hline
\end{tabular}
\end{table}

We verified the accuracy of the determined orientation in a~similar way.
Again, we placed the gate at random angles relative to the drone.
We collected the reference values manually using a~protractor.
The obtained results are presented in Table \ref{tab:rotation accuracy}.


\begin{table}[]
\centering
\caption{Selected test results for accuracy of orientation measurements in Euler angles.}
\label{tab:rotation accuracy}
\begin{tabular}{|c|c|c|c|c|c|c|c|c|}
\hline
\multicolumn{3}{|c|}{Camera rot. [$\deg$]} & \multicolumn{3}{c|}{Reference rot. [$\deg$]} & \multicolumn{3}{c|}{Absolute error [$\deg$]} \\
\hline
$\varphi$ & $\psi$ & $\theta$ & $\varphi$ & $\psi$ & $\theta$ & $\varphi$ & $\psi$ & $\theta$ \\
\hline
-178 & -27 & 89 & -180 & -25 & 90 & 2 & 2 & 1 \\
153 & 0 & 90 & 156 & 0 & 90 & 3 & 0 & 0 \\
-180 & 0 & 92 & -180 & 0 & 90 & 0 & 0 & 2 \\
177 & 2 & -180 & 180 & 0 & -180 & 3 & 2 & 0 \\
-148 & 7 & -1 & -149 & 5 & 0 & 1 & 2 & 1 \\
179 & 31 & -1 & 180 & 30 & 0 & 1 & 1 & 1 \\
171 & -1 & 1 & 180 & 0 & 0 & 9 & 1 & 1 \\
-171 & 3 & -2 & -180 & 0 & 0 & 9 & 3 & 2 \\
\hline
\multicolumn{6}{|l|}{Mean absolute error [$\deg$] (for all experiments)} & 3.14 & 1.41 & 1.18\\
\hline
\end{tabular}
\end{table}

The obtained results prove a~sufficient accuracy of determining both position and orientation with the used vision algorithm.
The average translation vector errors are significantly smaller than the size of the gate, so they do not have a~major impact on the flight efficiency.
The same is true for the orientation errors.
An angle difference of 3~degrees translates into about 40 mm of displacement at a~distance of 800 mm (value $d_1$ from the first strategy) -- far less than the gate size.


\subsection{Control accuracy}

We tested the performance of the control algorithms on a~track consisting of three randomly set gates.
For each described strategy, we conducted two series of eight flights: in daylight and in artificial light.
In doing so, we examined the efficiency of the flight through the gate, also in terms of the ''cleanliness'' of the flight (whether or not the vehicle has collided with the frame).
We present the obtained results in Table \ref{tab:control accuracy}.


\begin{table}[!t]
\centering
\caption{Test results of the effectiveness of both control strategies when flying through a~series of gates.}
\label{tab:control accuracy}
\begin{tabular}{|c|c|c|c|c|c|}
\hline
\multicolumn{6}{|c|}{1st control strategy} \\
\hline
\multicolumn{3}{|c|}{Artificial lighting} & \multicolumn{3}{c|}{Natural lighting} \\
\hline
No. & Passes & Collisions & No. & Passes & Collisions\\
\hline
1 & 3/3 & 1/3 & 1 & 3/3 & 0/3 \\
2 & 3/3 & 0/3 & 2 & 2/3 & 0/3 \\
3 & 3/3 & 0/3 & 3 & 3/3 & 0/3 \\
4 & 3/3 & 0/3 & 4 & 3/3 & 0/3 \\
5 & 3/3 & 0/3 & 5 & 2/3 & 0/3 \\
6 & 3/3 & 0/3 & 6 & 3/3 & 0/3 \\
7 & 3/3 & 0/3 & 7 & 3/3 & 0/3 \\
8 & 3/3 & 0/3 & 8 & 3/3 & 1/3 \\
\hline
OA & 100 \% & 4 \% & OA & 92 \% & 4 \% \\
\hline
\multicolumn{6}{|c|}{2nd control strategy} \\
\hline
1 & 3/3 & 0/3 & 1 & 3/3 & 1/3 \\
2 & 1/3 & 0/3 & 2 & 3/3 & 0/3 \\
3 & 3/3 & 0/3 & 3 & 3/3 & 0/3 \\
4 & 3/3 & 1/3 & 4 & 3/3 & 0/3 \\
5 & 2/3 & 0/3 & 5 & 3/3 & 0/3 \\
6 & 1/3 & 0/3 & 6 & 3/3 & 0/3 \\
7 & 3/3 & 1/3 & 7 & 3/3 & 0/3 \\
8 & 3/3 & 0/3 & 8 & 3/3 & 0/3 \\
\hline
OA & 79 \% & 8 \% & OA & 100 \% & 4 \% \\
\hline
\end{tabular}
\end{table}


The presented data shows that for the first control strategy, regardless of the way the room is lit, the flight efficiency is very high, between 90 and 100\%. 
Additionally, it is very rare for the drone to collide with the edge of the gate. 
For this method, the randomly checked flight time through a~series of three gates was about 26.5 seconds.


The second control strategy is characterised by a~significant drop in effectiveness under artificial lighting (down to 79\%).
In daylight, however, the flights are 100\% successful. 
For this strategy, the randomly checked flight time through a~series of three gates was 14.4 seconds, almost twice shorter than in the first solution.


As can be seen, the second control strategy allows a~significant increase in the speed of flight through the gates, but at the expense of efficiency in less favourable lighting conditions.
In our opinion, these errors are the result of DJI Trello's internal stabilisation algorithms.
Their accuracy depends on the quality of images recorded by the downward vision sensor.
Based on them, the current speed of the drone is calculated using the optical flow algorithm.
In the case of the second strategy, a~faster drone flight causes a~greater distortion of the images recorded by the downward vision sensor.
As a~result, artificial lighting conditions (in conjunction with the floor texture) are insufficient for the correct operation of optical flow, which decreases the accuracy of DJI Trello's internal stabilisation algorithms and therefore the effectiveness of realisation of the set trajectory.

\subsection{Implementation on embedded GPU}

On both the PC and the Nvidia Jetson TX2, we conducted performance tests of the designed system.
In the case of the eGPU platform, the function of visualising the drone's camera image along with the tagging of detected markers was disabled. 


When the algorithm was run on a~computer equipped with an Intel Core i5 processor and an Nvidia GTX 1060 graphics card, it achieved an average processing rate of around 40 fps.
For the GPU, it was around 25 fps. 
Both results are satisfactory and allow for successful flights.


\section{Conclusions}
\label{sec:conclusions}

The developed control system for the DJI Tello EDU drone proved to be successful.
Both presented strategies allow repetitive flights through a~track composed of gates with ArUco markers, similar to those used in the Alpha Pilot 2019 competition.
It is worth noting that they work only with vision feedback.
They do not need data fusion from multiple sensors (including IMU measurements) to generate a correct flight trajectory.
The application, implemented in Python, was run on a~latptop computer and on a~card with an embedded GPU.
In both cases, a~processing rate of at least 25 frames per second was achieved, which is sufficient to realise a smooth control.
The use of an embedded GPU will allow in the future, for a~different custom drone, to realise all processing in the vehicle's resources (on-board).


As part of our future research, we also plan to: conduct extensive tests in simulation (Gazebo with ArduPilot or PX4) and real-world, use AI-based gate detection (different types, not necessarily with ArUco markers), use more advanced control strategies and methods (data fusion, trajectory optimisation, approaches based on reinforcement learning), consider reconfigurable devices (FPGA, Zynq SoC) as the main computing platform.

\bibliographystyle{IEEEtran}
\bibliography{csk_mmar_2021.bib}


\end{document}